\documentclass[runningheads]{llncs}
\pdfoutput=1
\usepackage{cite}
\usepackage{amsmath,amssymb,amsfonts}

\usepackage{algorithm}
\usepackage{algpseudocode} 

\usepackage{graphicx}
\usepackage{textcomp}
\usepackage{xcolor}
\usepackage{subcaption}
\usepackage{sidecap}
\usepackage{lastpage}

\usepackage{fancyhdr}

\usepackage{booktabs}
\usepackage{multirow}

\def\BibTeX{{\rm B\kern-.05em{\sc i\kern-.025em b}\kern-.08em
    T\kern-.1667em\lower.7ex\hbox{E}\kern-.125emX}}

\begin{document}
\cfoot{\thepage\ of \pageref{LastPage}}

\title{TRACE: A Differentiable Approach to Line-level Stroke Recovery for Offline Handwritten Text}
\titlerunning{TRACE}
\author{Taylor Archibald\inst{1}\orcidID{0000-0003-2576-208X} \and
Mason Poggemann\inst{1} \and
Aaron Chan \inst{1}\orcidID{0000-0002-4022-5306} \and
Tony Martinez\inst{1}
}
\authorrunning{Taylor Archibald et al.}

\institute{Brigham Young University, Provo, UT\\
\email{taylor.archibald@byu.edu, martinez@cs.byu.edu}\\
\url{https://axon.cs.byu.edu}
}

\maketitle
\global\csname @topnum\endcsname 0 % NO FIGURES IN FIRST COLUMN!
\global\csname @botnum\endcsname 0

\begin{abstract}
Stroke order and velocity are helpful features in the fields of signature verification, handwriting recognition, and handwriting synthesis. Recovering these features from offline handwritten text is a challenging and well-studied problem. We propose a new model called TRACE (Trajectory Recovery by an Adaptively-trained Convolutional Encoder). TRACE is a differentiable approach that uses a convolutional recurrent neural network (CRNN) to infer temporal stroke information from long lines of offline handwritten text with many characters and dynamic time warping (DTW) to align predictions and ground truth points. TRACE is perhaps the first system to be trained end-to-end on entire lines of text of arbitrary width and does not require the use of dynamic exemplars. Moreover, the system does not require images to undergo any pre-processing, nor do the predictions require any post-processing. Consequently, the recovered trajectory is differentiable and can be used as a loss function for other tasks, including synthesizing offline handwritten text. 

We demonstrate that temporal stroke information recovered by TRACE from offline data can be used for handwriting synthesis and establish the first benchmarks for a stroke trajectory recovery system trained on the IAM online handwriting dataset.


\end{abstract}

\section{Introduction}
\begin{table}[t]
  \begin{tabular}{ll}
  (1) & \parbox[c]{1em}{
        \includegraphics[width=4.5in,trim={.01in 0 0 0},clip]{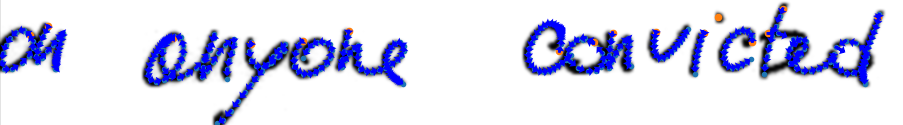}} \\
      (2) & \parbox[c]{1em}{
        \includegraphics[width=4.5in,trim={0 0 2.85in 0},clip]{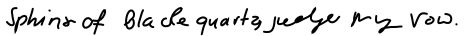}} \\
    
  \end{tabular}

\captionof{figure}{TRACE recovery and synthesis. (1) is a visualization of strokes recovered by TRACE from an offline handwriting image. Blue arrows indicate the predicted direction and orange points indicate the beginning of a new stroke. (2) is an example of a synthetically generated image that mimics the style of (1), and demonstrates how strokes recovered from offline data can be used for other tasks.}
\label{demo}

\end{table}
Handwriting is prevalent in both the physical and digital world. When handwriting is captured by a digital device, such as a pen-based computer screen, it is referred to as \textit{online} handwriting data. At a minimum, these data include the location of the pen tip or stylus when touching the screen through time~\cite{Plamondon:2000:OOH:331097.331275}. On the other hand, \textit{offline} handwriting data refers to digital images of handwriting inscribed on some physical medium.

While online data can be readily rendered as an image, the reverse process is much more difficult, as offline data lack a temporal component and often contain artifacts inherent to the writing medium or digitization process. Consequently, online handwriting data can make many tasks easier or more accurate~\cite{Plamondon1999},
including handwriting recognition, signature verification, writer identification, and handwriting synthesis (see Figure~\ref{demo}). While capturing handwriting online is becoming increasingly common, offline handwriting data collection can be easier in many instances and offline handwriting recognition remains an important challenge.

We propose a novel, differentiable model for stroke recovery called TRACE (\textbf{T}rajectory \textbf{R}ecovery by an \textbf{A}daptively-trained \textbf{C}onvolutional \textbf{E}ncoder). Our model is based on a CRNN that outputs a series of predicted stroke points, the number of which is proportional to the width of the original image. These predictions are then aligned using a Dynamic Time Warping algorithm (DTW) and compared against the ground truth (GT) stroke points to calculate a loss, from which the network is updated. The most important contribution TRACE offers is it extends prior trajectory recovery deep learning approaches to work on arbitrarily long lines of text. We provide the first trajectory recovery benchmarks for the IAM online handwriting database (IAM-On)~\cite{Liwicki2005} and the IAM offline handwriting database (IAM-Off)~\cite{marti2002iam}. Additionally, we demonstrate that strokes recovered by TRACE can be used to synthesize handwriting in the manner of a given style.

\section{Related Work}
Traditional, explicit methods to stroke recovery broadly split the problem into two phases: \textit{local examination}, where strokes are analyzed for startpoints, endpoints, loops, ambiguous zones, and other complications, and \textit{global reconstruction}, where the strokes are reconstructed based on features and observations derived from local examination~\cite{Nguyen2010}. A drawback of these approaches is that they often require handcrafted rules for each script. Moreover, they often rely on image preprocessing, including image skeletonization, which is sensitive to noise~\cite{Nguyen2010} and discards data that may be useful for determining the trajectory. With the advent of deep learning however, new approaches to trajectory recovery offer possible resolutions to these challenges.

Bhunia et al.~\cite{Kumarbhunia2018HandwritingNetwork} were the first to publish an end-to-end deep learning model for stroke recovery. Using an encoder-decoder style network, Bhunia et al. employed a CNN followed by an LSTM~\cite{hochreiter1997long}. Benefits of using an LSTM include that it can have an arbitrary number of states and can learn long-term dependencies. They demonstrate its effectiveness on single stroke characters with square images, although encoding the image as a finite length vector and their chosen loss function limit the applicability of this model to more difficult tasks, including wider, multistroke images. Hung Tuan Nguyen et. al. extend this approach to recover multiple strokes in a single image and add an attention mechanism in~\cite{nguyen2021online}, though they only demonstrate it for single Japanese kanji characters.

In~\cite{Zhao2018}, Zhao et al. use a CNN and dynamic energy prediction network for Chinese single-character recognition. In~\cite{Sumi2019}, Sumi et al. use a Cross Variational Autoencoder to translate from offline to online characters and vice versa. The process involves learning a shared latent space between online and offline representations of characters, by iteratively passing online data to one encoder and offline data to another encoder. Representations encoded in this latent space are then decoded into both an online and offline data representation, where the difference between the reconstruction and ground truth is used to tune the network, although it is only demonstrated for single characters. 

However, these efforts have focused on trajectory recovery for single characters and require significant revisions to process variable width images consisting of many characters.

\section{Method}
Our method broadly extends the ideas in ~\cite{Kumarbhunia2018HandwritingNetwork}, but we propose two additional methods to better align the prediction points with ground truth points, while also modifying the architecture to support encoding arbitrarily wide images, predict when a new stroke begins, and predict relative coordinates.

\subsection{Loss}
Our loss is broadly the distance between a sequence $T$ of target $(x,y)$ coordinates and the sequence of our predictions $P$ according to some mapping.

One goal of our loss function is it should favor stroke point combinations that could have generated the original image. This means that all predicted stroke points should lie somewhere on the original strokes, and they should collectively cover the entirety of the original strokes. We also desire for the model to accurately predict both the order the strokes were originally written as well as the direction of each stroke.

In~\cite{Kumarbhunia2018HandwritingNetwork}, the authors employed an $L_1$ loss, or the Manhattan distance from the predicted stroke points to the ground-truth points, where each predicted point is mapped to the GT point with the same index in the sequence. A potential issue with this approach is a set of stroke points that accurately reconstructs the original image might still incur a very large loss if the intervals between stroke points do not align well with the GT. Moreover, if the stroke has a small loop, the network might learn to exploit the loss by tending to place stroke points in the middle of the loop to minimize distance to all possible points in the loop if it cannot infer the direction of the stroke. 

Our goal is then to find a better mapping from the predicted points and GTs. Since many different sequences of points can define the same function, we do not constrain our investigation to a bijective mapping, matching each point in our prediction sequence to precisely one target. Rather, we consider many to many mappings that favor ensuring every predicted point is near a GT point, every GT point is near a predicted point, and the order of the predicted points mirrors that of the GT points. 

Formally, for all $t \in T$, we wish to minimize the distance between $t$ and the nearest $p \in P$, $p^*$. The constraint, 
\begin{equation}
\min_{p \in P} || t_i - p ||,
\end{equation}
ensures that our prediction spans the entire GT function $r$.

Similarly, we wish to minimize the distance between any $p \in P$ and the nearest point $t^* \in T$, i.e., 

\begin{equation}
\min_{t \in T} || p_i - t ||.
\end{equation}
This ensures that each $p$ is proximate to the original function, or that each predicted point lies on the original stroke. 

Finally, we wish to find a mapping that preserves the order each point appears within the stroke. Specifically, for $i$ indexing the sequence of points in some stroke and $s$ indexing some mapping, our mapping should require monotonicity

\begin{equation}
i_{s-1} \leq i_{s}
\end{equation}
and continuity
\begin{equation}
i_{s} - i_{s-1} \leq 1.
\end{equation}

A fast, dynamic programming algorithm that satisfies these constraints is dynamic time warping (DTW)~\cite{sakoe1978dynamic}. DTW is robust to translations and dilations along the ``time" dimension. DTW is typically computed by first computing a cumulative cost matrix, where each cell is the cumulative minimum cost needed to reach that cell. The last cell then provides the cost of the optimal alignment, while the optimal alignment can be solved with backward induction. To further reduce temporal complexity, we employ a uniform warping window on the cost matrix.

For our experiments, we prefer the $L_1$ loss, since $L_2$ penalizes outliers more and tends to produce more conservative predictions that fail to cover the corresponding GT strokes (e.g., it may predict only points in the middle of a crossbar in the letter ``t" ). However, a solution that predicts stroke points that span the entire crossbar but in the wrong direction might be a preferable for many tasks. 

\subsection{Adaptive Ground Truth}

In many instances, it is impossible to infer from the image the direction a stroke was drawn, or the order in which the strokes were drawn. A writer may cross a ``t with a left-to-right or right-to-left stroke. Similarly, a writer may cross a ``t or dot an ``i" immediately or upon the completion of a word or sentence. 

This presents a challenge for deep learning systems, since the loss function provides its most salient feedback if the system correctly predicts the order and direction of each stroke. However, for many applications, it is more important for the system to predict strokes that reproduce the image of the stroke in high fidelity, while being more invariant to the direction and order of the original strokes. 
A potential weakness of employing DTW as our mapping algorithm is that it enforces continuity and monotonicity on the entire sequence of strokes. While monotonicity and continuity should be enforced within each stroke, we wish to relax this requirement for a set of strokes to ensure that pathological strokes do not inhibit the ability of the system to make predictions that faithfully reconstruct the image.

Since we wish to preserve sequences of strokes while simultaneously minimizing the number of pathological model updates induced by reversed or out-of-order strokes, we employ a method for permuting stroke order and direction during training. Because each instance will be assessed by the model many times, we can adopt an iterative stroke reordering approach. Specifically, for each update, we perform at most one alteration to the GT strokes sequence, either swapping the order of adjacent strokes or inverting the sequence of points of a single stroke. We can identify candidate strokes by those with the greatest total or average DTW loss:

\begin{equation}
\Delta(P,T) := [\delta(P_i,T_j)]_{ij} \in \mathbb{R} ^ {n\times m}.
\end{equation}

To select a candidate to swap, we perform a softmax on this loss for each stroke to compute the sample probability of that stroke being altered:
\begin{equation}
\text{Softmax}(x_{i}) = \frac{\exp(x_i)}{\sum_j \exp(x_j)}.
\end{equation}

After sampling a stroke, we compute what the loss would have been had the GT stroke been altered by one of three transformations (if applicable): reversing the direction of the stroke, swapping that stroke with the next stroke, or swapping that stroke with the previous stroke. After performing this transformation, we recompute a new alignment and cost for this stroke. The stroke that yielded the lowest loss is saved as the GT for subsequent epochs, and the loss is computed relative to this adapted GT. As the number of iterations increases and each training instance is trained on multiple times, the model converges to a solution that better reconstructs the original images.

To improve efficiency, we only recompute the DTW alignment for some window around the affected stroke, and we do not need to recompute every possible cost, as the cost matrix prior to the affected stroke has not changed. Because we desire to preserve the actual stroke ordering and directions as much as possible, we employ the adaptive GT method to fine-tune the network after it has largely converged.

Even with these modifications, the temporal complexity of our loss function is still $O(N*k)$, where $N$ is the number of points and $k$ is the window size. To improve performance during training, the model can also be pretrained using a smaller DTW window and fine-tuned later with larger windows.

\subsection{Encoder-Decoder Network}

\begin{figure}
  \centering
  \includegraphics[width=.6\columnwidth]{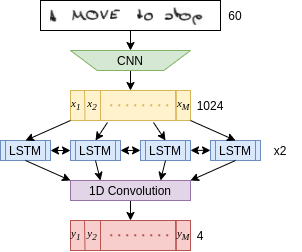}
  \caption{The network architecture for TRACE. The input image for the CNN is 60 pixels high with an arbitrary width. The resulting feature maps are approximately the same width as the input and 1024 feature maps. This is passed into a 2-layer, bi-directional LSTM, followed by a 1D convolution. The result is a sequence of stroke point predictions. }
  \label{AttentionModel}
\end{figure}

We employ a CRNN of the variety commonly used in handwriting recognition, depicted in Figure~\ref{AttentionModel}. Unlike ~\cite{Kumarbhunia2018HandwritingNetwork},~\cite{Zhao2018}, and~\cite{Sumi2019}, our model uses a variable-length encoder, which allows it to perform robust predictions for potentially arbitrarily wide images.

We start with an 11-layer CNN that expects the input images to be to 60 pixels tall, 1 channel, and any possible width, to handle handwriting segments of varying length. The output is a matrix of variable width, rescaled proportionally to the length of the width of the original image and 1024 feature maps. We use $3 \times 3$ kernels for convolution, and both $2\times2$ and $2 \times 1$ windows for MaxPool operations. %The model was generally robust to several common CNN architectures we tried.

\begin{figure}
  \centering
  \includegraphics[width=1.0\columnwidth]{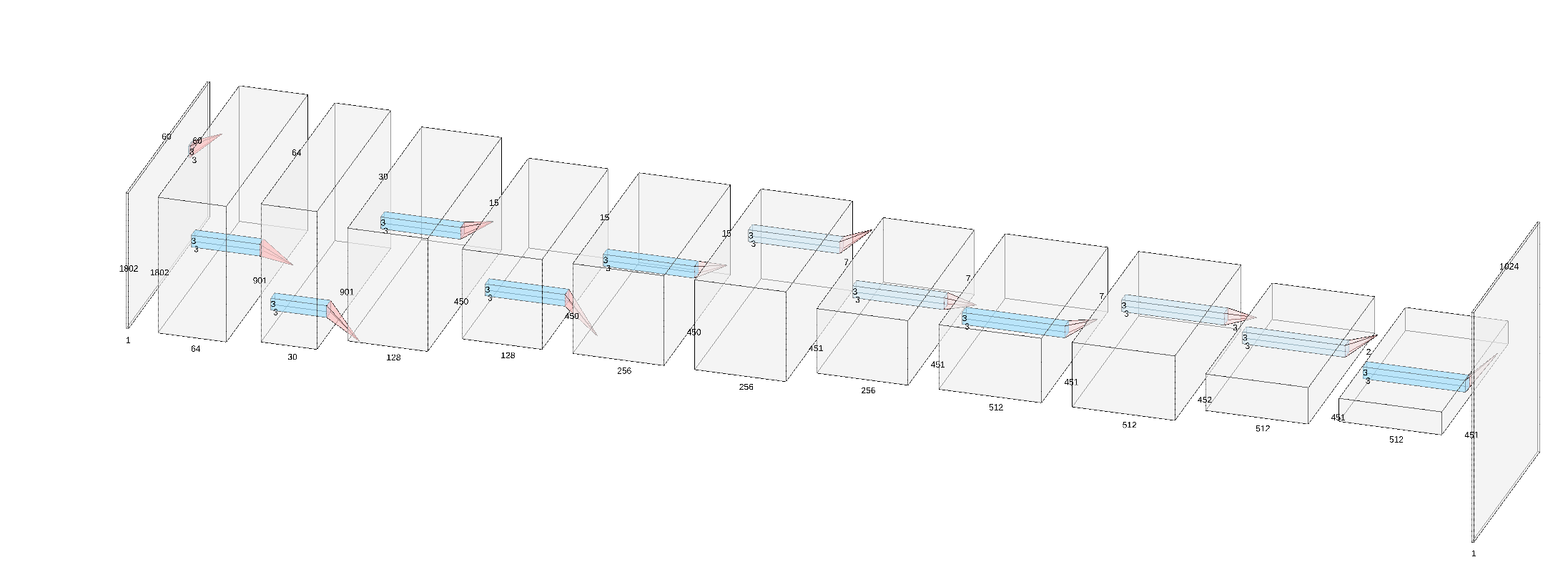}
  \caption{Our CNN architecture. We use primarily $3\times3$ kernels to perform the series of layer operations: Conv, MaxPool, Conv, MaxPool, Conv, Conv, MaxPool, Conv, Conv, MaxPool, Conv. The output is a matrix of variable width and a height of 1024.}
  \label{CNN}
\end{figure}

For each step, the CRNN predicts a relative coordinate $(x,y)$ from the last position, whether the point is the start of a new stroke (SOS token), and whether the point represents the end of a sequence (EOS token).

If the model is trained to predict relative coordinates with no other constraints, the resulting strokes tend to be reasonably accurate in isolation. However, collectively, these strokes often do not align well with the input image. This is due, in part, because pen-up movements are infrequent, and thus more difficult to learn, but simultaneously have a disproportionately large role ensuring the prediction is aligned to the original image. To achieve a kind of global consistency, we compute the cumulative sum of these relative coordinate predictions. We then employ an $L_1$ loss to compute the difference between these summed predictions and the GT absolute coordinates.

For predicting SOS tokens, we first compute the DTW alignment between the predicted stroke and GT. Once the alignment is computed, we consider the first predicted point that matches to a GT SOS point to be the corresponding SOS (i.e., when multiple predictions match to a GT SOS). We employ a cross-entropy loss with class weights due to the class imbalance between SOS and non-SOS points.

We similarly employ a cross-entropy loss when predicting the EOS. To mitigate the class imbalance issue for EOS tokens, we duplicate the EOS GT stroke point 20 times and append it to the end of the GT stroke sequence. The model is thus trained to predict an EOS token for all successive points after the first EOS stroke point (in contrast to the SOS process described above). This approach allows for the model to learn a smoother transition from non-EOS points to EOS points, while also mitigating the class imbalance issue.

For training, we used the ADAM optimizer~\cite{Kingma2015} with a batch size of 32, a learning rate of 0.0001, and a learning rate schedule that decreased the learning rate at a rate of .96 every 180,000 training instances.

\subsection{Data}

Since our goal is to reconstruct offline handwritten strokes, ideally we would have a set of offline images with corresponding online GT data. However, since these data are comparatively more difficult to collect, we adopt an approach of approximating offline data by rendering online data as images and degrading them. For our experiments, we train our model on the training and validation sets from the IAM online handwriting database (IAM-On)~\cite{Liwicki2005}, a corpus of trajectory data collected from 221 different writers and 10,426 lines. To validate the model, we test it on both IAM-On test set, as well as the entire IAM offline handwriting database (IAM-Off)~\cite{marti2002iam}, which is composed of 13,353 lines by 500 different writers.

Each line in IAM-On contains a series of positional coordinates $(x,y)$ as well as a time coordinate $t$. Since our model outputs predictions that are proportional to the width of the image, we resample the GT so that the number of stroke points is proportional to the width of the image. Moreover, because we are more concerned with recovering the shape of the strokes than the velocity, the points are resampled with respect to the cumulative stroke distance, so that points within a particular stroke are equidistant. While the model can be used to predict velocity as well by keeping the GT strokes parameterized by time, we prefer to use distance, as generally fewer points are needed to faithfully reconstitute the original strokes. Because of this reparameterization, TRACE is not technically recovering online handwriting data in our experiments (since it recovers this reparameterized data), though we have no reason to suspect TRACE would not work similarly well if it were trained on data parameterized by time.

Each set of strokes is then rendered as an image to be processed by the CNN. To better mimic offline data, a series of augmentations and degradations are applied to each image. These include varying stroke width and contrast, as well as applying random grid warping~\cite{Wigington2018DataNetwork}, random Gaussian noise, blurring, and other distortions~\cite{OCRtutorial}. We further supplement these data with 200,000 synthetic samples drawn from a generative model for online handwriting based on the one described in~\cite{graves2013generating} trained on the IAM-On training and validation data.

\section{Experiments}

We evaluate the success of our model using quantitative metrics for how well it recovers strokes from online data and offline data. We establish the first baseline performance for stroke recovery for lines of text on both the IAM-On and IAM-Off data. We also demonstrate how it can be used in online handwriting synthesis, a potential downstream task.

\begin{table*} 

\captionof{figure}{Random sample of IAM-offline stroke reconstructions} \label{offline_qual}
  \begin{tabular}{ll}
      (1) & \parbox[c]{1em}{
        \includegraphics[width=4.5in]{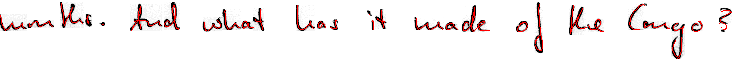}} \\
      (2) & \parbox[c]{1em}{
        \includegraphics[width=4.5in]{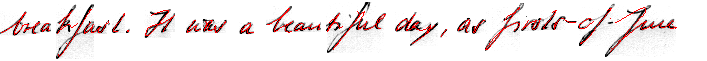}} \\
      (3) & \parbox[c]{1em}{
        \includegraphics[width=4.5in]{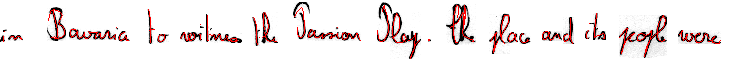}} \\        
      (4) & \parbox[c]{1em}{
        \includegraphics[width=4.5in]{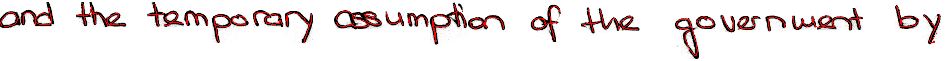}} \\
      (5) & \parbox[c]{1em}{
        \includegraphics[width=4.5in]{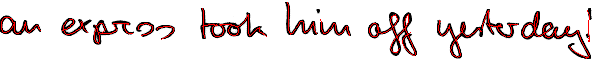}} \\
      (6) & \parbox[c]{1em}{
        \includegraphics[width=4.5in]{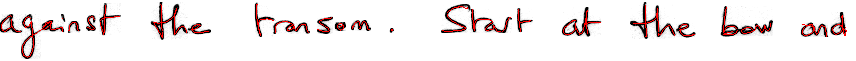}} \\
      (7) & \parbox[c]{1em}{
        \includegraphics[width=4.5in]{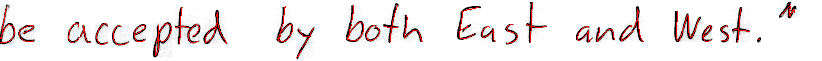}} \\
      (8) & \parbox[c]{1em}{
        \includegraphics[width=4.5in]{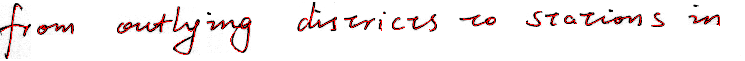}} \\
      (9) & \parbox[c]{1em}{
        \includegraphics[width=4.5in]{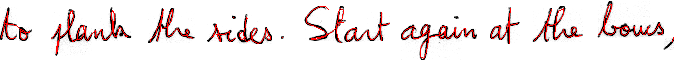}} \\

  \end{tabular}
  
\end{table*}

\subsection{Online evaluation}

\begin{table}
\begin{center}
\caption{Average DTW distance}
\label{table:dtw}
\begin{tabular}{|c|c|c|}
\hline
\multirow{2}{*}{Distance metric} & \multicolumn{2}{c|}{Average DTW loss}  \\
\cline{2-3}
& (equidistant GT) & (original GT) \\

\hline\hline
$L_1$ & 0.03060 &  0.03452 \\
\hline
$L_2$ & 0.02423 & 0.02745 \\
\hline
\end{tabular}
\end{center}
\end{table} 

We first consider how successfully the model is able to recover stroke trajectory information from online data rendered as images. Since the GT strokes are known, we use the DTW distance score between the GT stroke points and our predicted points (i.e., the cumulative sum of the relative points, as in the loss function). Table~\ref{table:dtw} reports $L_1$ and $L_2$ average DTW scores, both for the actual GT (where points are sampled as a function of time), as well as a resampled version where points are sampled as a function of cumulative stroke distance, as TRACE was trained to predict equidistant stroke points and ignore velocity. Additionally, because TRACE predicts more points than were in the original GT and predicting more points tends to decrease average DTW distances, we resample the predictions to have the same number of points as the original GTs. The DTW distances are scaled so that the distance from the lowest stroke point to the highest stroke point has a unit distance of 1.

% \parbox[c]{1em}{

\begin{table}
\begin{center}
\caption{Online NN distance}
\label{tab:nn_online}
\begin{tabular}{|c|c|c|}
\hline
\multirow{2}{*}{Type} & \multicolumn{2}{c|}{Average NN distance ($L_2$)}  \\
\cline{2-3}
& (equidistant GT) & (original GT) \\
 
\hline\hline
GT to nearest prediction & 0.01662 & 0.01751 \\
\hline
Prediction to nearest GT & 0.01405 & 0.01615  \\
\hline
\end{tabular}
\end{center}
\end{table} 

We also report the average distance to the nearest neighbor (NN distance). In this case, we measure the distance between each predicted point and the nearest GT point and vice versa. Measuring each predicted point to the nearest GT is a measure akin to precision, and measures the extent to which the predicted points lie somewhere on the GT. Conversely, measuring the distance between each GT point to the nearest prediction resembles recall, and measures how well the predictions cover the entire space of GT points. Table~\ref{tab:nn_online} suggests that TRACE is slightly better at ensuring the predicted stroke points are near GT stroke points than it is at ensuring every GT stroke point is near a predicted point, which is supported by the observation that TRACE does not dot every ``i'' or cross every ``t". As with DTW loss, using equidistant GT points decreases error.

\subsection{Offline evaluation}
Because IAM-Off does not have ground truth strokes, we cannot compute the DTW distance. Instead, we consider the average NN distance from each predicted point to the nearest GT pixel (as opposed to the nearest stroke point), and define a GT pixel on the image as one that has an intensity of less than 127.5 on a scale from 0 to 255, which creates many more GT points for comparison than in the online experiment. This result is reported in Table~\ref{tab:nn_offline}.

\begin{table}
\begin{center}
\caption{Offline NN distance}
\label{tab:nn_offline}
%\begin{tabular}{|c|p{1.5in}|}
\begin{tabular}{|c|c|}
\hline
Type & Average NN distance ($L_2$)  \\
\hline\hline
Prediction to nearest GT & $9.311 \times 10^{-4}$ \\
\hline
\end{tabular}
\end{center}
\end{table}

\begin{table}
\begin{center}
\caption{LPIPS metric}
\label{tab:lpips}
\begin{tabular}{|c|c|}
\hline
Model & LPIPS metric  \\
\hline\hline
White image (baseline) & 0.423 \\
\hline
TRACE without DTW & 0.296 \\
\hline
TRACE & \textbf{0.106} \\

\hline
\end{tabular}
\end{center}
\end{table}

In Table~\ref{tab:lpips}, we report the Learned Perceptual Image Patch Similarity (LPIPS)~\cite{zhang2018unreasonable} metric between the offline image and an image of the reconstructed strokes using a 2 pixel stroke width. Specifically, we report the Alex-lin variant LPIPS metric for TRACE, an identical model but trained without DTW loss, and a baseline of all white images.

Qualitative results of the process can be observed in Figure~\ref{offline_qual}, which shows a random sample of GT offline images with the recovered strokes overlaid in red. TRACE tends to do very well in predicting neat, well-spaced handwriting, and generally better when the handwriting stands in high contrast to the background. TRACE struggles somewhat with predicting punctuation, as well as isolated strokes, or strokes that are often not drawn consecutively with respect to the rest of the character (as in the dot in an ``i" or the cross in a ``t"). Moreover, it sometimes fails to accurately predict stroke extremities, or strokes that approach too near the top of the line (as in the ``P" in "Passion" on line 3). 

\begin{figure}
    \includegraphics[width=4.7in,trim={.5in 0 3in 0},clip]{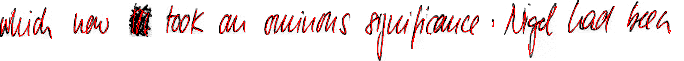}
    \captionof{figure}{TRACE achieves robust performance despite the presence of an anomalous marking.}
    \label{scribble}
\end{figure}

Figure~\ref{scribble} shows it is partially robust to anomalies as it successfully resumes after an aberrant, scribbled out marking.

\subsection{Synthesis evaluation}
While recovering handwritten stroke trajectory may be of interest for its own sake, often it is considered an intermediate step for solving other handwriting tasks, including handwriting recognition, handwriting synthesis, and writer verification. While there are many possible ways to incorporate recovered strokes into these systems, we demonstrate one way it can be done for handwriting synthesis.

One of the first handwriting synthesis models is described in~\cite{graves2013generating}. In this model, an LSTM is used to parameterize a Mixture Density Network (MDN), which can then be iteratively sampled to predict each successive stroke point. Data can then be passed to the LSTM to prime the model to mimic a particular handwriting style. 

The success of this particular model depends on the input data being structured as a sequence of points, rather than as an image, as in the case of offline handwriting. In our case, we use recovered strokes from offline data to variously train or prime the model to synthesize offline handwriting styles.

Using offline handwriting in this system is useful not only for cases when the target handwriting style has not been captured online, but also because online and offline handwriting styles are not the same. Factors that contribute to these differences include the friction between the writing surface and implement, the responsiveness and sensitivity of the digital screen, and any changes to the way a person holds his or her hand when writing on a digital surface (e.g., to not touch the screen with the side of his or her hand). Moreover, strokes recovered from offline text can be used as a way to augment the training data of these kinds of systems, which improves generalization.

Figure~\ref{synth_styles} demonstrates the ability of the synthesis system to mimic the style of an offline sample. Note that all synthetic texts are rendered with the same stroke width and consequently mimic only the rough shape of the original input and not, e.g., line quality. For synthesizing experiments, we synthesize English pangrams, sentences that include every letter of the alphabet at least once. Our test sentences include 

\begin{itemize}
  \item Sphinx of black quartz, judge my vow.
  \item The five boxing wizards jump quickly.
  \item How vexingly quick daft zebras jump.
\end{itemize}

As Figure~\ref{synth_styles} shows, more common letters and n-grams generally appear to produce better results. 

\begin{table}

  \begin{tabular}{ll}
     %\multicolumn{1}{c}{\textbf{text}} & \multicolumn{1}{c}{\textbf{text}} \\

      (a) & \parbox[c]{1em}{
        \includegraphics[width=4.5in]{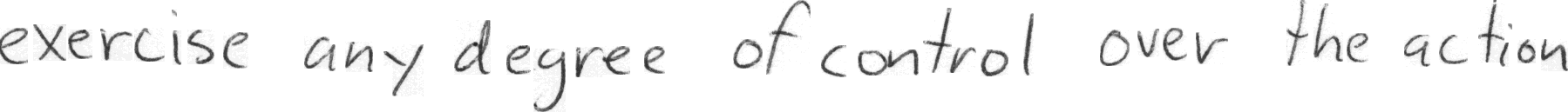}} \\
      & \parbox[c]{1em}{
        \includegraphics[width=4.5in]{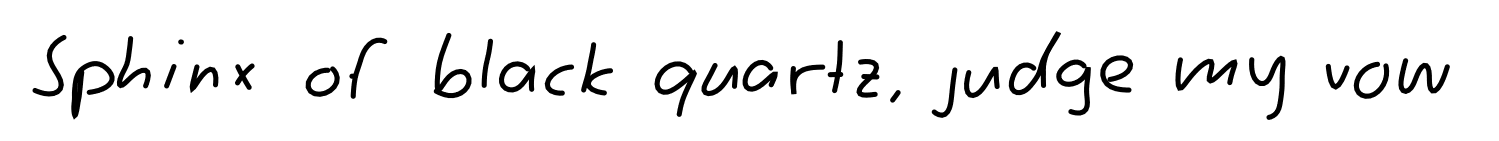}} \\
      (b) & \parbox[c]{1em}{
        \includegraphics[width=4.5in]{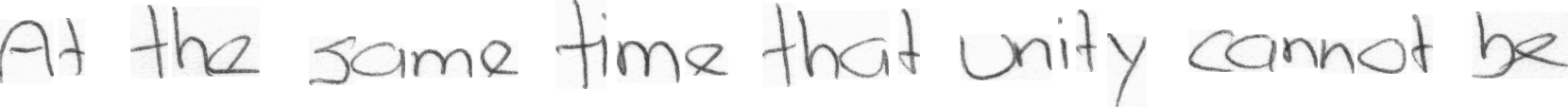}} \\        
       & \parbox[c]{1em}{
        \includegraphics[width=4.5in]{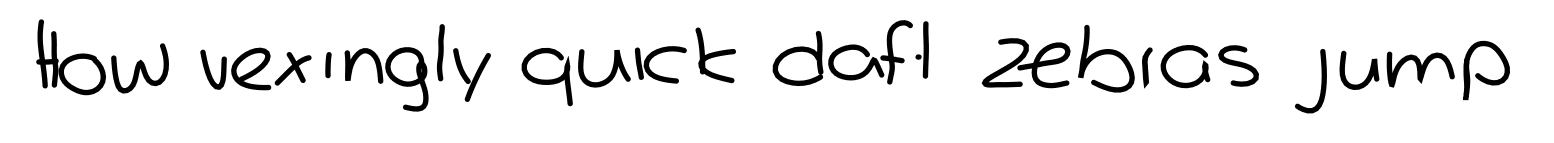}} \\
      (c) & \parbox[c]{1em}{
        \includegraphics[width=4.5in]{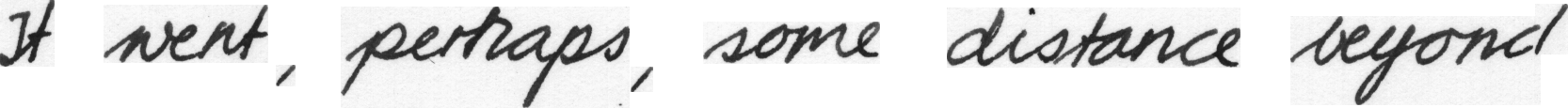}} \\
       & \parbox[c]{1em}{
        \includegraphics[width=4.5in]{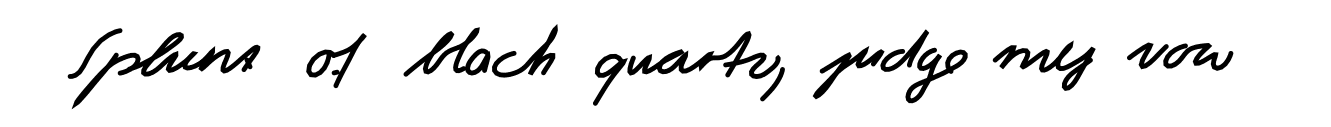}} \\
      (d) & \parbox[c]{1em}{
        \includegraphics[width=4.5in]{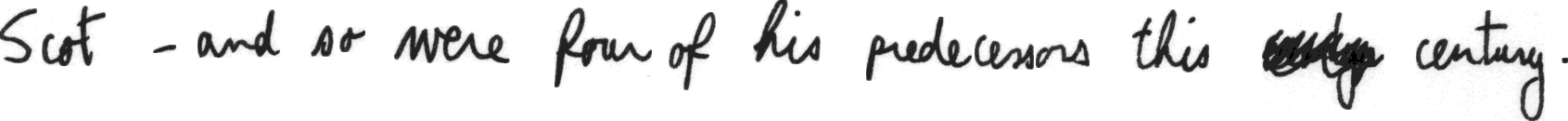}} \\
       & \parbox[c]{1em}{
        \includegraphics[width=4.5in]{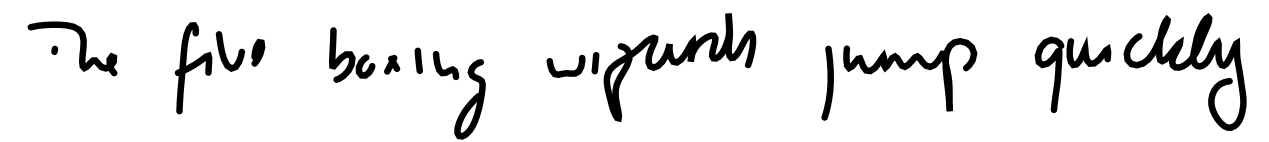}} \\

  \end{tabular}

\captionof{figure}{Each pair of lines above constitutes (1) an offline image used to prime the synthetic text model and (2) an image of synthetic text generated by the model in the style of the offline image.} \label{synth_styles}

\end{table}

\subsection{Synthetic Evaluation}
\begin{table}

  \begin{tabular}{ll}
     %\multicolumn{1}{c}{\textbf{text}} & \multicolumn{1}{c}{\textbf{text}} \\

      (a) & \parbox[c]{1em}{
        \includegraphics[width=4.5in]{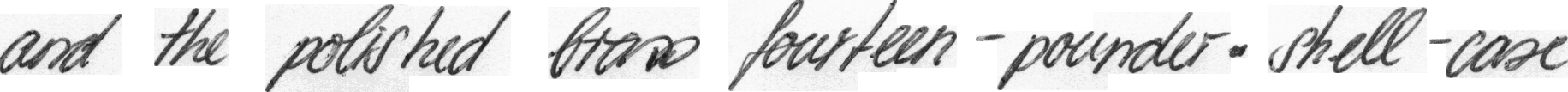}} \\
      (b) & \parbox[c]{1em}{
        \includegraphics[width=4.5in]{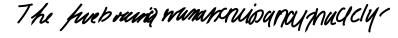}} \\
      (c) & \parbox[c]{1em}{
        \includegraphics[width=4.5in]{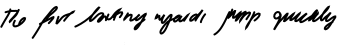}} \\
      (d) & \parbox[c]{1em}{
        \includegraphics[width=4.5in]{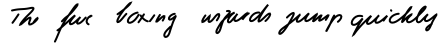}} \\

  \end{tabular}

\captionof{figure}{Comparison of synthesized text based on different training data. (a) is a sample of offline handwriting data used to prime the handwriting synthesis model. (b), (c), and (d) are results from models trained on only online data, only offline data, and both online and offline data, respectively. The synthetic text is the pangram ``The five boxing wizards jump quickly."} \label{synth_on_off_comparison}

\end{table}

Figure~\ref{synth_on_off_comparison} compares text generated by a system trained only with online data, another with only converted offline data, and another one trained on both, all being primed with a converted offline sample. While all three achieve various success in this task generally, the system is more prone to produce worse or degenerate samples when primed with a style it has not been trained on, as noted in~\cite{graves2013generating}.

Note that the inability of the synthesis system trained only on native online data to synthesize text when seeded with converted offline data may be due, in part, to either the imperfections that arise from the conversion process or inherent distinctions between online and offline handwriting styles. When the system is trained on offline data, and particularly when it is trained on other samples of a particular author, the system produces better synthetic samples. 

Thus, TRACE enables seeding the synthesis model with an offline sample to mimic an offline handwriting style, while also augmenting the set of training data available to the synthesis model, enabling it to produce better synthetic handwriting samples.

\begin{figure}
\centering
\begin{subfigure}{.5\textwidth}
    \centerline{\includegraphics[width=1.0\textwidth]{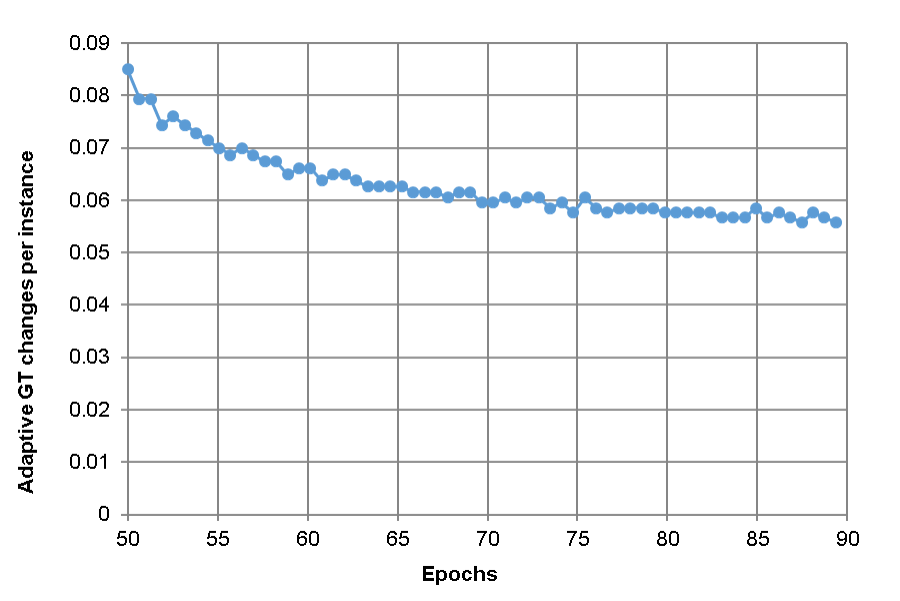}}
    \caption{The percentage of instances that undergo a GT adaptive change converges.}
    \label{ada_through_time}

\end{subfigure}%
\begin{subfigure}{.5\textwidth}
\centerline{\includegraphics[width=1.0\textwidth]{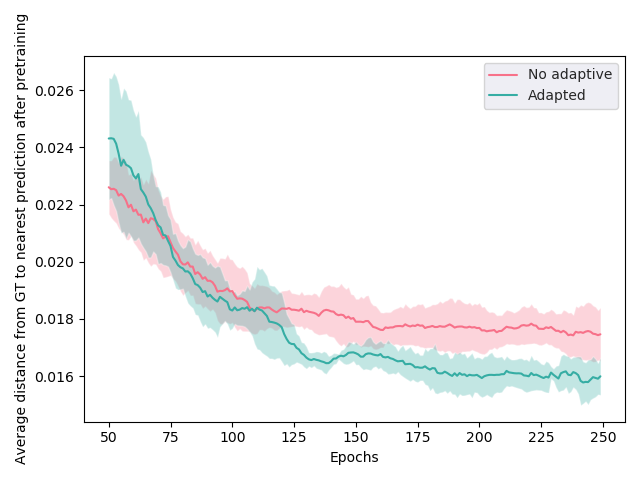}}
\caption{Adapting GTs on the training set improves NN loss on test set.}
\label{loss_through_time}
\end{subfigure}

\caption{Adaptive GT performance by epoch}
\label{fig:test}
\end{figure}

\subsection{Adaptive GT Ablation Study}
To demonstrate how the adaptive GT can improve the system, we first train the network for 50 epochs with the original GTs to have the system first learn the GT stroke orders and directions, before fine-tuning it with adaptive GTs to improve the system's ability to handle anomalous strokes.

Once the network has been pretrained, we employ the adaptive GT algorithm described in Section 3. Figure~\ref{ada_through_time} shows how the number of swaps and changes is small to begin with, as a change to the GT is helpful for fewer than 1 in 10 instances. After 40 additional epochs, the number of changes has started to converge with 35\% fewer changes per instance than initially.  Figure~\ref{loss_through_time} shows how the average NN loss on 5 runs  converges to a lower loss than without the use of adaptive GTs on the test set.

\section{Conclusion}

In this work, we have proposed TRACE, a novel method to recover stroke trajectories from offline data, and, in effect, rendering it in an online format. TRACE works well on wide images composed of many characters and strokes and is completely differentiable. We have demonstrated that it can be used to enable online handwriting synthesis to work with offline data.

A possible future application of TRACE is it can be used as a loss function for synthesizing an offline handwriting image directly (as opposed to synthesizing strokes only) and could supplement approaches presented in \cite{davis2020text} and \cite{fogel2020scrabblegan}. Another possibility would be using it to augment training data for online recognition systems. 

To improve the stroke recovery method, one direction might be to use a generative model, such as training a mixture density network as in~\cite{graves2013generating,Graves2016}, or an invertible neural network~\cite{Dinh2016, ardizzone2018analyzing}. This would afford the model greater ability to model uncertainty and provide multimodal solutions. Another possible method to improve performance is to reparameterize input strokes as Bézier curves, as in~\cite{Carbune2019}. 

\section{Acknowledgements}
We would like to thank Chris Tensmeyer for his suggestions and feedback.

%\section*{References}

%\bibliographystyle{ieeetr}
\bibliographystyle{splncs04}
\bibliography{bib/manual_bib, bib/proposal_ref}

\vspace{12pt}

\end{document}